\documentclass[preprint,12pt]{elsarticle}

\usepackage{graphicx}
\usepackage{amssymb,amsmath,hyperref}

\usepackage{lineno}
\usepackage{multirow}
\usepackage{makecell}
\usepackage{xcolor}

\journal{Computers and Geosciences}

\begin{document}

\newcommand{\ans}[1]{{\color{black}#1}}
\newcommand{\ansnew}[1]{{\color{black}#1}}

\begin{frontmatter}

\title{Hybrid and Automated Machine Learning Approaches for Oil Fields Development: the Case Study of Volve Field, North Sea}



\author{Nikolay O. Nikitin\corref{cor1}}
\cortext[cor1]{Corresponding author: nnikitin@itmo.ru}
\author{Ilia Revin, Alexander Hvatov, Pavel Vychuzhanin, Anna V. Kalyuzhnaya
\footnote{Nikolay Nikitin: methodology, software development, visualization, writing of original draft.  Ilia Revin:  software development, visualization, writing of original draft.  Alexander  Hvatov:   methodology,  writing  of  original  draft.   Pavel  Vychuzhanin: software development,  validation.   Anna  Kalyuzhnaya:  conceptualization,  project  administration.}}
\address{ITMO University, 49 Kronverksky Pr. St. Petersburg, 197101, Russian Federation}

\begin{abstract}
The field development workflow contains numerous tasks involving decision-making processes. The modern machine learning methods, including automatic machine learning (AutoML), reduce the geophysics or machine learning experts' time required to solve routine tasks. In the paper, we focus on the automated solution of the location of the wells optimization problem, namely, improving the quality of oil production estimation and estimating reservoir characteristics for appropriate wells placement and parametrization, using the same AutoML approach. Ideas of making several parallel or consequent tasks automatically within one framework are arising as Composite AI. We implemented and investigated the quality of forecasting models for oil production estimation: physics equation-based, pure data-driven, and hybrid. CRMIP (Capacitance-Resistance Model Injector-Producer) model is chosen as a physics-related approach. We automated the seismic analysis using evolutionary identification of convolutional neural network structure for reservoir detection to help investigate reservoir characteristics for wells location choice. The Volve oil field dataset was used as a case study to conduct the experiments. The implemented approaches can analyze different oil fields and even be adapted to similar physics-related problems.

\end{abstract}


\end{frontmatter}


\section{Introduction}

 Nowadays, a process of field development decision making includes modern technologies such as geological simulation for reservoir reconstruction, hydrodynamic reservoir simulation for oil production forecast, 3D visualization for interpretation of obtained results. 
 
\begin{figure}[ht!]
\centerline{\includegraphics[width=1.0\textwidth]{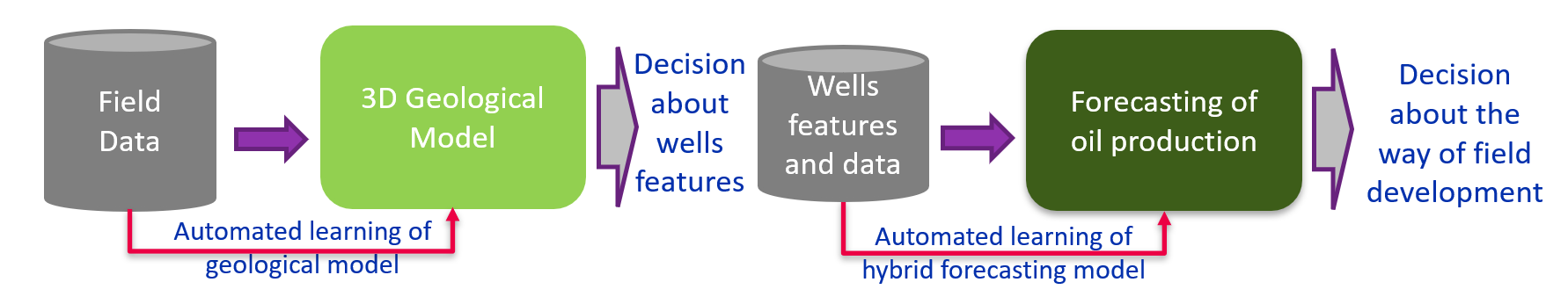}}
\caption{Decision making pipeline for field development \ans{using automated and hybrid modelling techniques}}
\label{fig:pipeline}
\end{figure}

 \ans{Nevertheless, the normal process of decision making in oil production companies requires a deep involvement of experts on each step of the pipeline to repeat the same routine procedures every time the new site appears. In the paper, we suggest an intelligent solution for automation of modeling steps (Fig.~\ref{fig:pipeline}).
} 

It is undeniable that many geophysical problems can be solved \ans{manually} using physics-based models. For example, liquid dynamics in porous media and hydrodynamic models are widely used to analyze and develop oil sites. Nevertheless, their use is often related to a series of issues: computational complexity of the implemented models and the long operating time and limitations that appeared in the sites with heterogeneous structure and physical properties followed by reducing the simulation results' accuracy. Therefore, the data-driven approaches usually balance the range of applicability of the model, the quality of the prediction, and the computational complexity.

\ans{Another example of the expert routine problem is the seismic exploration. In the paper, we consider seismic modeling as a sub-task of modeling oil fields, which is devoted to analyzing the reservoir. Careful analysis by a qualified expert determines the main stratigraphic units within the seismic volumes that may contain hydrocarbons. These patterns can be determined objectively and are considered predictable \citep{nanda2016seismic}.}

 Machine learning is widely known as an effective and useful instrument in geophysical modeling \citep{khatibi2020machine}. The well-developed techniques are \ans{already exist for different stages to help the modeling expert. As an example there are tools} for oil production forecasting \cite{sagheer2019time}, oil fields historical analysis \cite{li2019deep}, seismic analysis of reservoirs \cite{qiang2020prediction} and other problems \cite{hanga2019machine}. \ansnew{Nevertheless, the existing method exchanges the geophysicists' time with the machine learning expert's time to build and train the models. In the paper, we try to highlight the possibilities of machine learning as a treatment for the lack of intelligent automation for decision-making in the process of field development.}
 
 \ans{To make the scope more precise,} we focus \ans{mainly} on the particular problem of building efficient, intelligent automated methods for assistance in oil field development, more precisely:
 
\begin{itemize}
\item automated methods for recognition of reservoir characteristics (with seismic data) in regions surrounding wells for preliminary (before oil production estimation step) decision making;
\item automated methods for achieving high-quality forecasting of oil production with given parameters of wells (using pure machine learning, physics-related models, and hybrid models).
\end{itemize} 

\ans{In the seismic exploration area, the number of modern machine learning methods applications is small. However, we still can mark the process of stratigraphic units detection automation, as an example \citep{di2017multi}, which ultimately increases the chances of research success. It should be noted that for many tasks related to the analysis of seismic data, it is advisable to use standard methods for processing seismic data involving manual data labeling and manual model training for every particular site.}

\ans{As an example of successful routine automation} in the petroleum industry, capacitance-resistance models (CRM) tuning for the given site is widely used. The classical problem of the CRM model calibration is in the scope of the various classical \cite{sayarpour2009field}, and recent papers \cite{temizel2019improving,wang2019improved}. However, choosing between plenty of the existing history-matching approaches is required to obtain the desired quality for the given site. There are many approaches \cite{holanda2018state}, including the Gradient projection method within a Bayesian inversion framework, SQP, BFGS, GAMS/CONOPT, StoSAG, various genetic algorithms, ensemble Kalman filter.

The hybrid (composite) approach \cite{kansao2017waterflood,temirchev2019reduced} can be used instead of pure data-driven \cite{zhao2015insim,artun2016characterizing,esmaili2016full} methods to improve a single CRM model's performance. In the hybrid approach, a combination of the liquid dynamics model and the machine learning models and adaptation methods and modeling automation methods are being studied to improve the characteristics of tools for modeling, forecasting, and optimizing field development. Therefore, often CRM models are combined with machine learning techniques. The pure data-driven approach excludes the liquid dynamics of physics-based models, leading to non-realistic predictions.

\ans{Theoretically, we can use inversion algorithm \cite{evensen2007using} for the forecasting oil production  using numerical reservoir simulator (e.g. ECLIPSE \cite{pettersen2006basics}). Moreover, a numerical reservoir simulator may be used within a hybrid model. It allows obtaining reliable long-term forecasts of the oil production rate without human supervision. However, history matching is computationally expensive since many simulation runs are required within the optimization algorithm. }

\ans{All of the approaches described above require either expert in the geosciences or data scientist time. To reduce the time expenses,} we propose the approach that allows us to combine different types of models and optimize them using automated modeling techniques. \ans{Such automation may not lead to the most precise prediction. However, it still gives reasonable prediction with lesser human effort}. We achieve it with the FEDOT framework \cite{nikitin2022automated}, which enables the possibility of automatic ensembling of the models that fit best to the given historical data. With a little effort to wrap the CRM model (i.e., implementing an adapter for the existing CRM model), the framework automatically builds the hybrid and the pure data-driven model. 

\ans{The most preferred option for seismic exploration is the use of automatic machine learning methods that take on the task of interpreting large volumes of seismic data and allow experts to use time-saving automation methods. Determining the characteristics of a reservoir usually requires the involvement of experts. However, machine learning models can also solve seismic analysis problems formulated as image recognition problems. In this case, a convolutional neural network can analyze seismic slices. However, the question arises about determining the optimal structure of these networks. In the article, we propose to use an evolutionary approach to solve it.}

Overall, the unified approach allows the building of the reservoir and the other data-driven models as soon as the data are available. Moreover, \ansnew{it is done in the same way}, and, thus, several models could be obtained. In the paper, two interconnected cases are considered: machine learning is applied to reservoir modeling and seismic exploration problem featuring the Volve dataset \cite{VolveDS}.

The paper is organized as follows. Sec.~\ref{sec_problem} provides the formulation of the oil field analysis and modeling problems that was chosen to solve in the article. Sec.~\ref{sec_data_and_models} described the dataset \ans{that} was used as a case study and the basic models that were applied to process it. Sec.~\ref{sec_hybrid} describes the details of the application of the automated evolutionary search of hybrid models for oil production forecasting. Sec.~\ref{sec_seismic} describes the proposed oil field seismic analysis approach based \ans{at the automated neural architecture search}. Sec.~\ref{sec_exps} contains the description of the experimental studies, and Sec.~\ref{sec_concl} summarizes the obtained results and provides its brief analysis.

\section{Problem statement}
\label{sec_problem}

The reliable analysis and modeling of the oil field development \ansnew{cannot} be achieved without applying the mathematical models. The classical approaches of the oil field modeling in machine learning are based on the material balance reservoir models (usually CRM and their modifications). However, obtaining high-quality results from the balance models without the direct involvement of the domain expert is complicated since the expert selects the appropriate history matching and data preprocessing strategies. 

For these reasons, we decided to propose a data-driven approach for the oil field analysis. The data-driven oil production forecasting task can be formalised as:

\begin{equation}
\begin{aligned}
{} & q_i(t), q_i(t+1), ..., q_i(t+f) = \\
& F(Q(t-1), {I}(t-1), {Q}(t-2), {I}(t-2), ..., , {Q}(t-w), {I}(t-w))
\end{aligned}
\end{equation}

where $q_i$ - production for the producer $i$, $Q$ - productions for \ans{each} producers, $I$ - injections for the each injector, $t$ - time step, $w$ - size of the historical time window, $f$ - forecast length, $F$ - function of data-driven model.

The identification of the data-driven model structure representing $F$ with the experts' minimal involvement can be achieved using the evolutionary approaches for structural learning. The widely used approach for this task is automated machine learning (AutoML). The $F$ can be evaluated using the pure-ML and hybrid (composite) approaches that combine CRM and ML advantages.

Therefore, we propose investigating the efficiency of the different machine learning approaches to solve the described problem in the following ways:

\begin{itemize}
\item  Implement the data-driven solution based on a single ML model;

\item  Implement the data-driven solution based on a hybrid model that combined ML and partially physics-related CRM;

\item  Replace the single ML model to the composite model \ansnew{consisting of} several blocks;

\item  Apply evolutionary optimization to find the most effective structure of the composite hybrid model;

\item  Analyze the efficiency of the proposed methods against the baseline (CRM model).
\end{itemize}

For the seismic analysis of the oil field, a different formulation is required. We can get seismic time slices during the prepossessing of registered seismic records (seismic wavefields). In a conventional seismic processing workflow, the reflected portion of the wavefield is enhanced at the expense of the rest of the wavefield \cite{difraction2021}. By analyzing these sections, we can determine the spatial position of seismic boundaries, where the reflection of seismic waves occurs. These boundaries usually have a specific lithological and stratigraphic reference in the geological section. Using this particular pattern, we determine the corresponding features of the studied geological section based on the data of seismic studies \cite{yilmaz2001seismic}. In favorable geological and geophysical situations, it is possible to directly indicate hydrocarbon deposits' location.

Also, one of the essential stages of interpretation of seismic survey data is the search for the contact boundary between two layers of different geological elements of a rock mass with different physical properties. This boundary is usually called the horizon. In order to find such horizons, it is usually used to analyze seismic sections (in image format) of a seismic cube with the help of an expert geologist. This analysis aims to search for patterns in the images of the seismic sections to highlight various geological elements of a rock mass. 

Since seismic data interpretation is complex and requires expert knowledge, it is very relevant to conduct direct searches for hydrocarbon deposits using a seismic data-driven method. This issue has been widely discussed in seismic exploration for many years. Direct detection of hydrocarbon deposits by registered amplitude anomalies is possible only when the recording equipment and systems obtain records in actual ("raw") amplitudes. 

In particular, abnormally high negative values of reflection coefficients between the gas-saturated reservoir and the overlying clay rocks give rise to intense amplitude anomalies. For the seismic traces, such anomalies in the amplitudes of reflected waves are commonly referred to as "bright spots" \cite{hilterman2001seismic}. The boundary between the gas and the other part of the reservoir filled with oil or water appears on seismic time slices in the form of a flat subhorizontal line. The recording of such a reflection is called a "flat spot" in practice. In some cases, a direct indicator of hydrocarbon deposits' presence is the appearance of a so-called "dark spot" - the area with no regular recording. This feature of seismic recording is explained by the low impedance of gas-saturated sands or limestones and the same low impedance of clay tires.

Thus, the task of the direct search for hydrocarbon deposits using seismic data can be reduced to classifying two-dimensional representations of the seismic time slices. In its basic form, this is a binary classification problem. The negative class includes slices without any anomalies in the amplitudes of reflected waves meaning that potential hydrocarbon deposits are less likely to appear in such zones. The positive class includes slices containing any type of anomalies in the amplitudes of reflected waves (any kind of "spots")\cite{faultdetection2019}.

Finding the contact boundary between two geological elements can be reduced to the problem of binary semantic segmentation of two-dimensional representations of seismic time slices. We have images of the seismic time slice and corresponding masks obtained with the help of experts. Masks contain information about the location of the contact boundary on a given seismic time slice.

We represent a seismic time slice $S$ as a combination of two sets $Z_{inline}^{T}\cup Z_{crossline}$. $Z_{inline}$ and $Z_{crossline}$ contain seismic tracks in a time-series format in the depth direction and across the seismic section, respectively. Since we use time axes perpendicular to the slice section and directed to the depth of this slice, for $Z_{inline}^{T}$, we use a transposed form of notation. From a mathematical point of view, it can be formulated, as shown in Eq.~\ref{eq:time_slice}.

\begin{equation}
\label{eq:time_slice}
Z_{crossline}, Z_{inline}^{T}=[(z_{1}),... ,(z_{T})]
\end{equation}

In Eq.~\ref{eq:time_slice} each vector $z_{i}$ contains time series of recorded seismic data at the specific receiver $i$.

We assume that the set of pairs "object, class" $X\times Y$ is a probability space with an unknown probability measure $P=P(S_{i}|y_i), \, S_{i}\in X, \, y_i \in Y$. There is a finite training sample $D \in X\times Y$ (Eq.~\ref{eq:sample_space}) of observations generated according to the probability measure. 

\begin{equation}
\label{eq:sample_space}
D =[(S_{1},y_{1}),... ,(S_{m},y_{m})]
\end{equation}

We need to build an algorithm that estimates probability measure $P$ and thus build the function 
$f:X\to Y= \text{arg} \max \limits_i P(S_{j}|y_i)$ that can classify an arbitrary object $S_j \in X$.

\section{Data and models for oil field forecasting}
\label{sec_data_and_models}

We use the Volve field as a case study for the oil field analysis and modeling. The Volve field was discovered in 1993. It is located in the central part of the North Sea, 5 km north of Sleipner East field, water-depth 80 m. The reservoir in the Volve field consists of Jurassic sandstones at 2750 – 3120 m. The Volve open dataset \cite{VolveDS} is the comprehensive oil site dataset that contains both oil production and seismic records. For the CRM model, we take the "production data" part. For seismic modeling, we take the "ST0202ZDC12.POSTSTACK.FULL" part. The main characteristics of both datasets are shown in Tab.~\ref{tab:volve_data}.

\begin{table}[h!]
\resizebox{\textwidth}{!}{%
\centering
\begin{tabular}{|c|c|c|}
\hline
Dataset characteristic                                                            & Variable type                   & Value                                                                                                                  \\\hline
Temporal range                                                                    & \multirow{9}{*}{Oil production} & 01.09.2007-17.09.2016                                                                                                  \\\cline{1-1}\cline{3-3}
Temporal resolution (days)                                                        &                                 & 1                                                                                                                      \\\cline{1-1}\cline{3-3}
Number of wells                                                                   &                                 & 7                                                                                                                      \\\cline{1-1}\cline{3-3}
\multirow{6}{*}{Used measurements}                                                &                                 & averaged downhole pressure                                                                                             \\\cline{3-3}
                                                                                  &                                 & averaged downhole temperature                                                                                          \\\cline{3-3}
                                                                                  &                                 & oil volume                                                                                                             \\\cline{3-3}
                                                                                  &                                 & gas volume                                                                                                             \\\cline{3-3}
                                                                                  &                                 & water volume                                                                                                           \\\cline{3-3}
                                                                                  &                                 & \begin{tabular}[c]{@{}c@{}}coordinates from IRAP RMS \\ model configuration \\ provided within the dataset\end{tabular} \\\hline

In-line range                                                                     & \multirow{7}{*}{Seismic}        & 9961 - 10361 - 1                                                                                                       \\\cline{1-1}\cline{3-3}
Cross-line range                                                                  &                                 & 1961 - 2600 - 1                                                                                                        \\\cline{1-1}\cline{3-3}
Z range (ms)                                                                      &                                 & 4 - 4500 - 4                                                                                                           \\\cline{1-1}\cline{3-3}
\begin{tabular}[c]{@{}c@{}}Inline and Crossline bin size \\ (m/line)\end{tabular} &                                 & 12.50/12.50                                                                                                            \\\cline{1-1}\cline{3-3}
Area (sq km)                                                                      &                                 & 40.10                                                                                                                  \\\cline{1-1}\cline{3-3}
Survey type                                                                       &                                 & Both 2D and 3D                                                                                                         \\\cline{1-1}\cline{3-3}
Used measurements                                                                 &                                 & Poststacked seismic traces                      \\
\hline
   
\end{tabular}}
\caption{Oil production and seismic characteristics of the Volve field used for the experimental setup implementation.}
\label{tab:volve_data}
\end{table}

\begin{figure}[ht!]
\centerline{\includegraphics[width=1.0\textwidth]{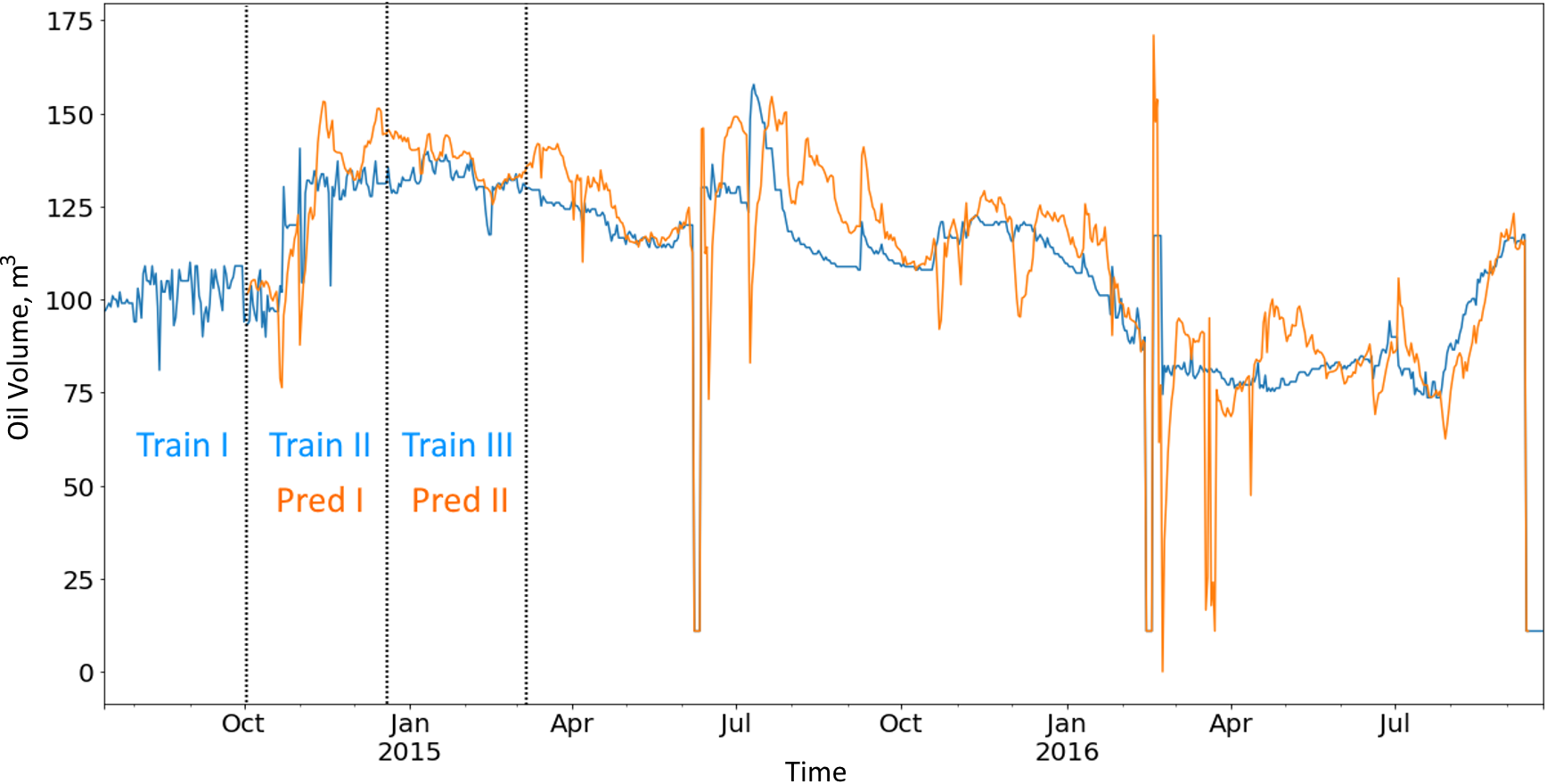}}
\caption{The strategy of the window-based optimization of the CRMIP model parameters (history matching). The blue line represents the observed oil production data, the orange line the prediction.}
\label{fig:temporal_window}
\end{figure}

Among the different implementations of the CRM \cite{holanda2018state}, we choose the CRMIP \cite{yousef2006capacitance} as the most suitable for the available data. Overall the hybrid approach described below is not restricted to the type of CRM. However, we assume that the particular model's history matching procedure is implemented. For CRMIP in the article, a gradient descent method for the history matching is chosen. In the experiments, we use the SLSQP optimization method implemented in the SciPy package \cite{virtanen2020scipy}. Additionally, we extended the usual history matching procedure with a temporal window optimization strategy. The main idea of the strategy is illustrated in Fig.~\ref{fig:temporal_window}. 

In Fig.~\ref{fig:temporal_window}, the scheme of the model application is provided. The first temporal window "Train I" is used for gradient descent history matching. After that prediction, "Pred I" uses the resulting parameters. Then the temporal window "Train II" is used for history matching. The process loops over all the available train windows. The resulting combined predictions are considered the prediction of the CRMIP model and compared with the historical oil volume.
We vary window length to assess the uncertainty of the prediction. In particular, we estimate the probability intervals of the oil volume debit for a given timestamp using all predictions obtained for different window lengths.

\begin{figure}[ht!]
\centerline{\includegraphics[width=1.0\textwidth]{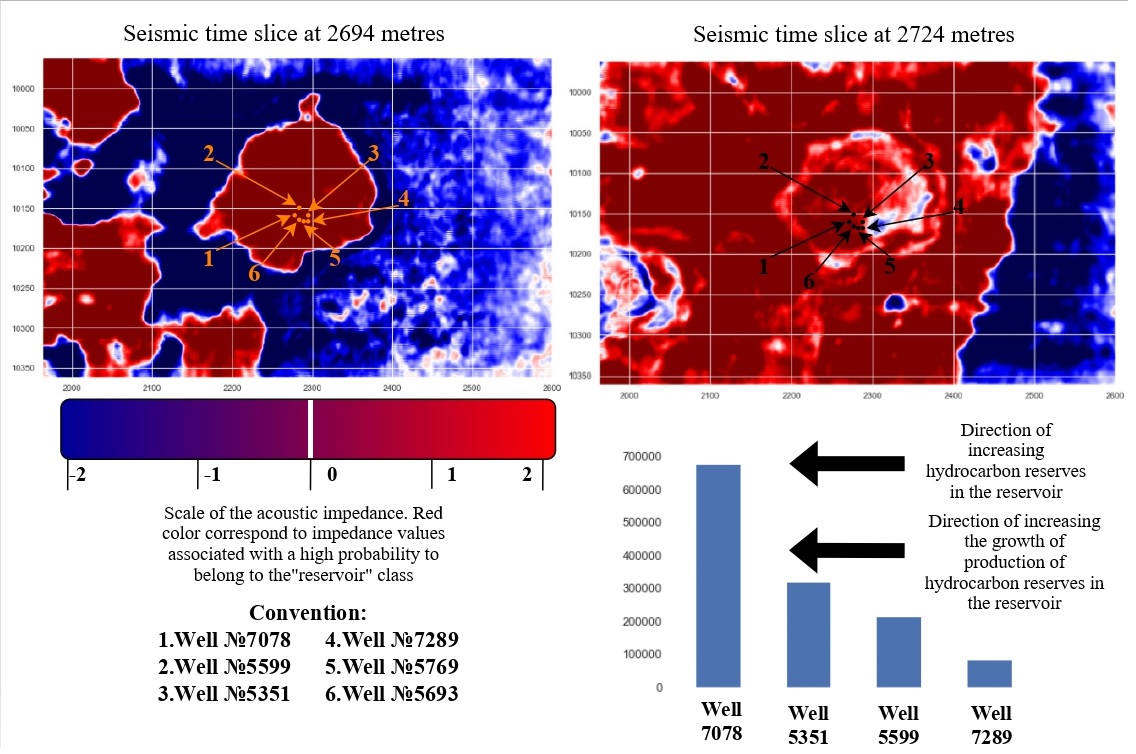}}
\caption{Combination of seismic analysis and oil production estimations. The seismic time slice (in the right) shows that oil production in the wells increases in the direction of oil reservoir.}
\label{fig_seismic_and_production_combination}
\end{figure}

In Fig.~\ref{fig_seismic_and_production_combination}, we can see how the oil production of the well and the direction in which the oil reservoir extends are interconnected. The left seismic time slice shows the top of the reservoir.

\section{Hybrid modelling}
\label{sec_hybrid}

The modern machine learning approaches make it possible to solve real-world problems from different fields (including oil production forecasting \cite{liu2020forecasting} and other physics-based \ansnew{\cite{krasnopolsky2020using,de_Figueiredo_2021}} problems) with high efficiency. \ans{The data-driven forecasts can be used to produce both long-term and short-term forecasts \cite{tadjer2021machine} and predict the future performance of the reservoir \cite{jeong2018learning}. The involvement of additional preprocessing allows handling even non-stationary time series \cite{sun2019building}}.

At the same time, the oil field analysis and modeling refer to the class of long-known problems with a well-developed mathematical foundation (that is noted in Sec.~\ref{sec_data_and_models}). It can be used to improve the quality of such forecasts. For many physics-related fields (e.g., climate \cite{krasnopolsky2006complex}, river floods forecasting \ansnew{\cite{WaterMdpi}}, microseismic events monitoring \ansnew{\cite{Zhu_2021}}, classification and prediction of rock types \ansnew{\cite{3Dgeo2021}} and ocean \ansnew{\cite{o2018integrated,ocean2021,spectrum2020}} modeling), the application hybrid models that combine the data-driven and physics-based components allow obtaining better results in comparison with the stand-alone components. \ans{The other issue in the production forecasting is the estimation of the uncertainty \cite{jeong2018cost}. There are specific \ansnew{approaches to} the robust model design \ans{existing} that allow preserving the appropriate quality even under high uncertainty - the examples can be provided for both oil production forecasting \cite{zhong2021deep} and other geo-related fields \cite{vychuzhanin2019robust}.}

The other forecasting technique that can be used for oil field forecasting is ensemble modeling \cite{helmy2013non}. The AutoML-based approaches can be used to identify the optimal structure of the ensemble or hybrid model or to solve specific geophysics tasks \ansnew{\cite{curvesclusters2021}}. In the paper, we use the open-source automated modeling and machine learning framework Fedot\footnote{https://github.com/nccr-itmo/FEDOT}. The framework allows applying the evolutionary structural learning to find the best suitable structure of the composite (hybrid) model using crossover, mutation, selection, and regularisation operators for the models represented as a directed acyclic graph \cite{kalyuzhnaya2020automatic}.

\begin{figure}[ht!]
\centerline{\includegraphics[width=1.0\textwidth]{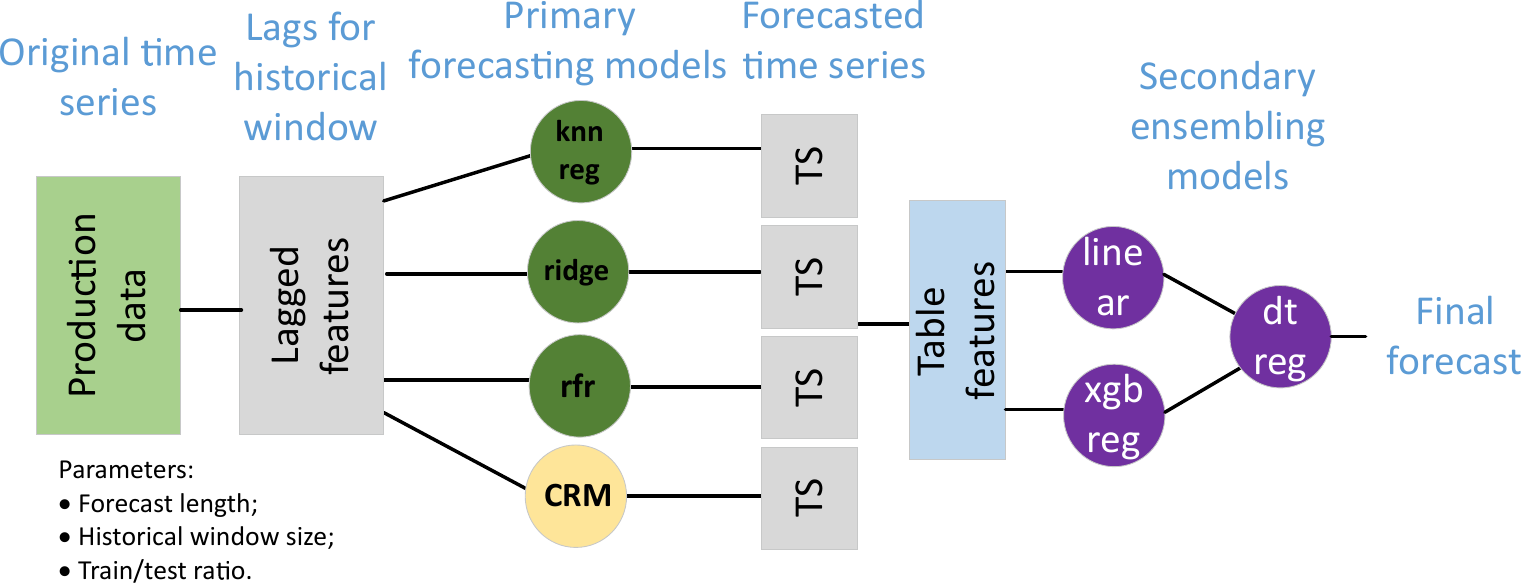}}
\caption{The structure of the hybrid composite model for oil production forecasting. The CRM block is the implementation Capacitance Resistance Model (in the paper the CRMIP modification is used). The green and violet blocks represent the possible candidates for machine learning regression models \ansnew{based on} k-nearest, ridge, random forest, linear, decision tree, XGBoost, LSTM, etc. The additional task-based hyperparameters of the composite model are forecast length and historical window size.}
\label{fig_composite}
\end{figure}

The possible structure of the hybrid composite model that includes both CRM and machine learning models is presented in Fig.~\ref{fig_composite}. This model can be used to solve the oil production forecasting problem.

We represent it as a multivariate time series forecasting task. Many strategies exist that can be used to improve the quality of results \cite{bontempi2012machine}. In a frame of composite model implementation, the set of data transformations is used. We emphasize two main types of transformation: the lagged transformation of the oil production history and the ensembling of the multi-model forecasts. \ans{So, we use the history of production for each well, using well log data from the Volve Field \ansnew{\cite{logdata2021}} as a component of a multivariate time series. Then, the lagged transformation is applied to obtain a trajectory matrix used as a features table for regression models. The prediction of CRM is used as additional exogenous features.}

The set of the experimental studies is conducted (the results are available in Sec.~\ref{subsec_exp_prod_forecast}) to investigate the efficiency of such approaches for the Volve dataset. \ans{Their main aim is not to overcome all existing approaches to the oil fields modeling (which is impossible due to the model and data limitations) but to prove the correctness and effectiveness of the proposed automated hybrid approach for the real-world case against widely used CRM-based approach.}

\section{Seismic analysis}
\label{sec_seismic}

The Volve subsurface dataset included full and partial angle stacks of two seismic surveys (ST0202 and ST10010), which have been co-processed for time-lapse analysis through a TTI Kirchhoff pre-stack depth migration (PSDM) workflow. The wells' ties confirm the data is close to zero phases, and a peak in seismic amplitude corresponds to an increase in acoustic impedance \citep{volvo4D2020}. \ans{The idea of classification is to search for geological elements with similar acoustic impedance values. Moreover, although the acoustic impedance values of the chalk formation and Hugin sandstone reservoir are similar, the geometric shape of these elements allows them to be quite clearly distinguished from each other. Thus, we avoid false-positive cases of detection of the reservoir while forming a sufficient volume of the training data set}. The second approach that we use is the application of deep neural networks to solve the problem of seismic time slices classification. The purpose of the experiment is to classify seismic data into those that can potentially contain an oil reservoir and those that do not. Usually, the inversion and regularization methods are used to solve this problem \cite{cooke2010model}. \ans{There are various algorithms for implementing seismic inversion, for example, such as geostatistical seismic inversion \ansnew{\citep{geotensor2018}}, but they all gives the possibility to build static reservoir models for field development and estimate hydrocarbon availability}. 

In the paper, we propose an approach based on seismic time slices before inversion transformation. We solve a binary classification problem for seismic slices interpreted as images. To assign an image of a seismic slice to any of the classes, we use seismic slices in the same period but after inversion processing. With the help of an expert, the image is assigned to one of two classes: "Potential oil reservoir" or "No oil reservoir". The proposed approach allows us to get information about potential oil reservoirs, bypassing the stage of inversion processing and expert evaluation.

\subsection{Seismic data preprocessing.}

\begin{figure}[ht!]
\centerline{\includegraphics[width=1.0\textwidth]{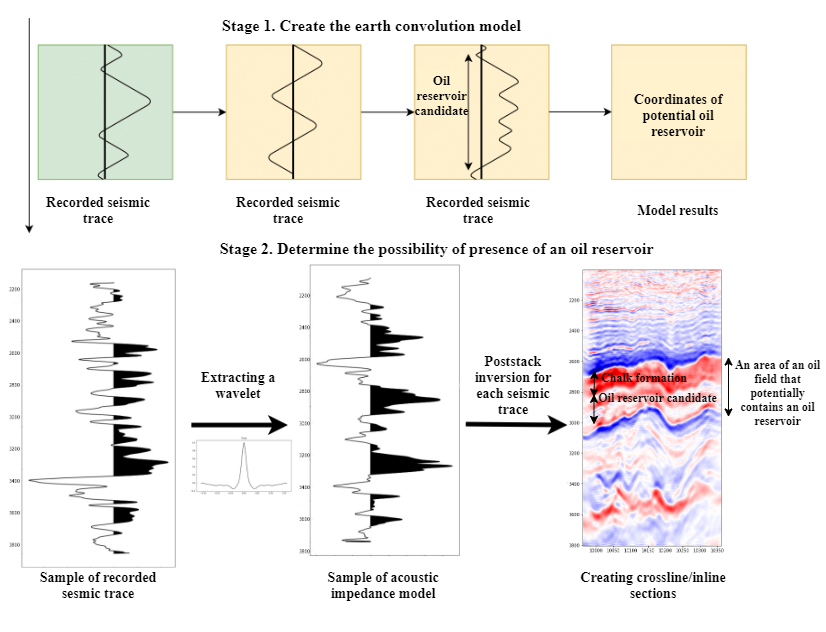}}
\caption{Flow chart of seismic data preprocessing approach for Volve seismic data. As can be seen from the figure, the key point is to interpret values of acoustic model impedance on the resulting seismic sections. Detecting the presence of a potential oil reservoir requires cross-comparison of results of the acoustic impedance model and seismic sections.}
\label{fig_Volve_inversion}
\end{figure}

\ans{To transform the raw seismic trace, we have applied an approach consisting of two stages. Stage 1 is to apply the well-known approach to building an earth convolution model. Stage 2 is to determine the possibility of the presence of an oil reservoir. We need to analyze the post-stack inversion model using all available sections (inlines, crosslines) and get more precise boundaries of geological elements and potential oil reservoirs. The result of this approach is the coordinates of the location of the potential oil reservoir. This information can be used as a supplement when analyzing the volume of hydrocarbon reserves and evaluating the field from the point of view of its profitability}. A flowchart for this approach is shown in Fig.~\ref{fig_Volve_inversion}.




\subsection{Model data preprocessing.}

\begin{figure}[ht!]
\centerline{\includegraphics[width=1.0\textwidth]{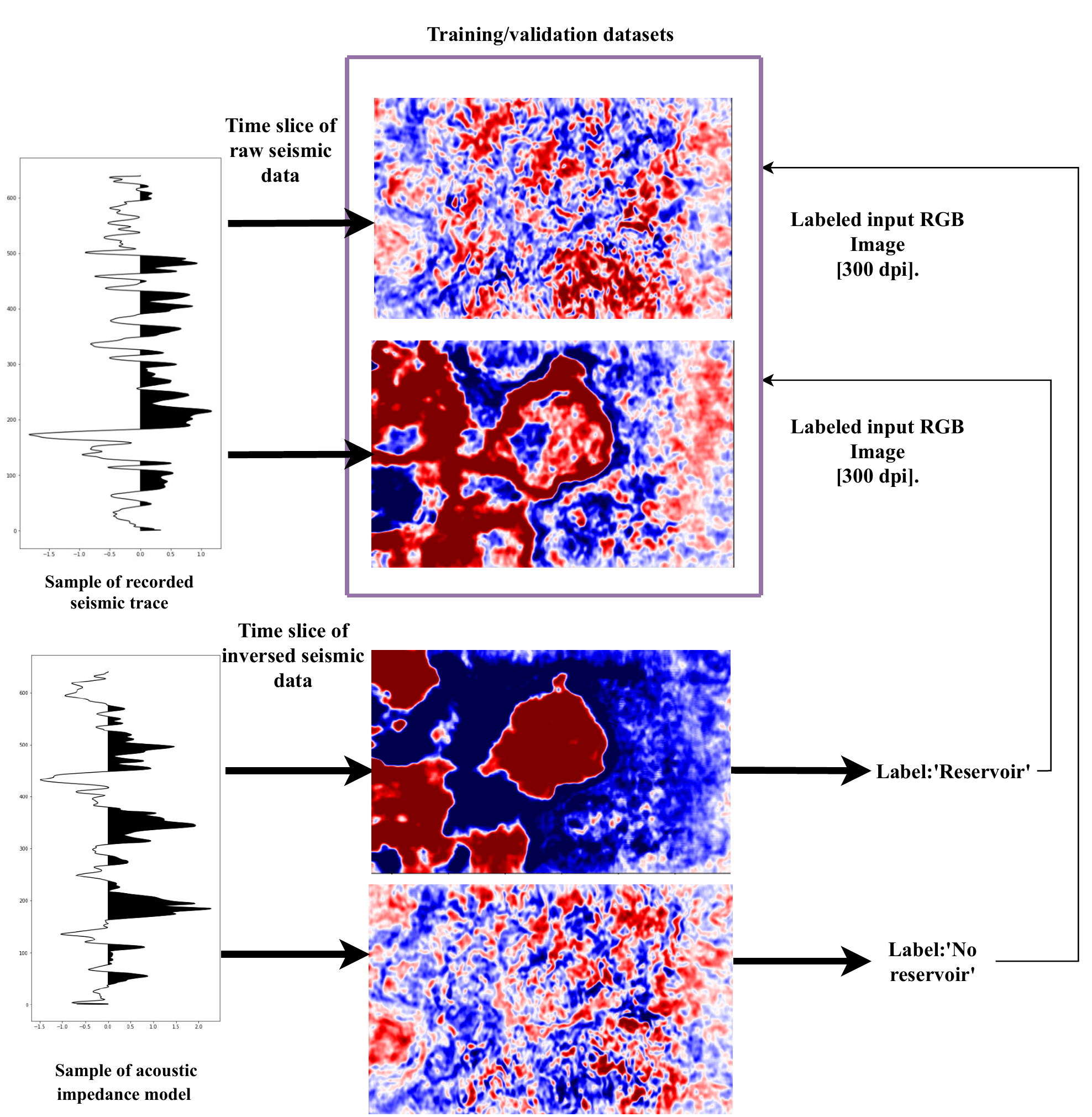}}
\caption{Flowchart for the generation of training and test data sets, based on seismic time slices images. After using the proposed approach to preprocessing seismic data, the resulting seismic time slices are divided into two classes ('Reservoir' and 'No Reservoir'). Then label of the inverse seismic time slice is assigned to the original seismic time slice.}
\label{fig_train_and_test_data}
\end{figure}

 The appropriate seismic data sample should be prepared to implement the data-driven model to solve the described reservoir detection problem. As a result of inversion processing, we obtained 450 images of seismic time slices. After expert analysis, training, validation, and test data sets were formed, including 290, 130, and 30 samples, respectively. The samples were divided into two classes in half to avoid the problem of unbalanced classes. A pipeline of the seismic data preprocessing is shown in Fig.~\ref{fig_train_and_test_data}.

From a mathematical perspective, the image of the seismic time slice can be represented as a function $I: N^2 \to N $ of two variables $i$,$j$.

\begin{equation}
\label{eq:image_function}
image := I(i,j) \to k
\end{equation}

Function $I$ generally defined in a rectangular shape, but for the convenience of research work, all images are determined by the square areas $i=0,1,...,W$, $j=0,1,...,W$ , $0 \le k \le 255,  k \in N$ where $W$ is image width, $H$ - height of the image and $k$ variable is responsible for the color ``intesity''.

The resulting data set based on the Volve dataset is too small to solve image classification. Unfortunately, this is a prevalent and realistic case: data collection can sometimes be extremely expensive or almost impossible in many applications. Thus, in addition to building the model, we have two more tasks: 

\begin{itemize}
\item Creating artificial samples for our data set;
\item Searching for the near-optimal neural network architecture for this task.
\end{itemize}

We use two techniques to expand our training data set with new samples to create new images. The first technique is aimed to make copies of the original images by rotating them around the axis to a certain degree. The second technique involves applying various filters to the original or converted images, which introduces noise and distortion into the images, so the model never sees twice the same picture. This technique prevents overfitting and helps to generalize the model.

\subsection{Deep learning for the seismic slices classification.}

Several types of machine learning algorithms that link acoustic impedance models and post-stack seismic data are used to solve the reservoir detection problem in a data-driven way. The resulting ML-based model allows us to connect the distribution of properties of the oil reservoir in the field. 

These tasks can be reduced to image classification tasks or object detection tasks. If we consider the classification problem, it can be divided into binary and multi-class classifications. The choice of task type depends on the research goals. We can either predict the absence/presence of an oil reservoir in a seismic section or predict a specific "class" of an oil reservoir based on the description given in Section~\ref{sec_problem}. 

The goal of an object recognition problem is to determine the oil reservoir's location, size, and properties. Examples of using machine learning and deep learning algorithms for seismic inversion problems are shown in \cite{LI2019inversion,qiang2020prediction}.

We consider the seismic slices as 2D images. Therefore the model should classify these images in a robust and precise way. The practical approach for such a problem is the application of the convolutional neural networks \cite{rawat2017deep}.

The manual selection of network architecture for this machine learning task requires expert knowledge in the applied field and neural networks. Thus, we use the neural architecture search (NAS) to find the optimal neural network architecture.

In order to select the best suitable architecture of the CNN-based model reservoir detection without the direct involvement of the expert, we used the extension of the FEDOT framework - the evolutionary NAS tool \footnote{https://github.com/ITMO-NSS-team/nas-fedot} that allows identifying the optimal architecture of the convolutional neutral classifier \cite{mamontov2018evolutionary}. 
\ans{Since reservoir detection is an extremely complex and nonlinear problem that depends on many factors, we decided to use a probabilistic estimation of whether the selected seismic time slice (in the case of binary classification) or pixel (in the case of semantic segmentation) belongs to the "reservoir" class. If the probability of being assigned to a class exceeds 80 percent, the selected object is classified as part of the reservoir.}

\section{Experimental studies}
\label{sec_exps}

The set experiments are conducted to analyze the efficiency of the ML-based methods for the oil fields modeling and analysis. There are two groups of experiments that can be distinguished: oil production forecasting and automated seismic analysis.

\subsection{Oil production forecasting}
\label{subsec_exp_prod_forecast}

The experimental studies devoted to data-driven oil production forecasting are provided using the following setup. The implementation of the CRMIP model was used for the partially physics-related forecasting, and the set of ML models is used to construct the data-driven model. 

\ans{The aim of this experiment is to prove the applicability of hybrid and automated approaches described in Sec~\ref{sec_hybrid} for the obtaining of effective forecasting models for oil production in an automated way. The main issue here is the relatively small historical data sample for the Volve field.}

\ans{The forecast lengths between 100 and 400 days were used during the experiments. The reason for the selection of relatively short forecast length is the limited time range of available production data for Volve (2-3 years). Also, the CRM models usually do not perform well on long-term forecasts with daily scale \cite{kansao2017waterflood}. At the same time, the application of the AutoML requires an extensive training sample for the appropriate model design.}

\begin{figure}[ht!]
\centerline{\includegraphics[width=1.0\textwidth]{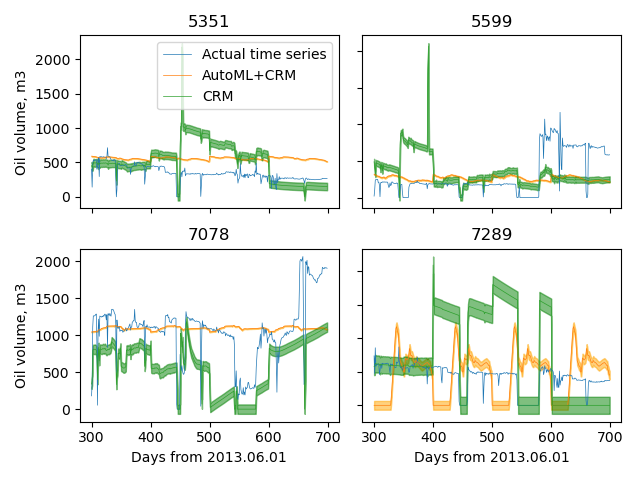}}
\caption{\ans{The comparison of the mid-term forecasting quality for the different approaches modelling: CRM-only model and AutoML-based data-driven pipeline. The evolutionary optimisation was applied to identify the optimal structure of the hybrid model. The forecast length is 400 days.}}
\label{fig_prod_forecasting}
\end{figure}

The forecasting results obtained with different approaches are presented in Fig.~\ref{fig_prod_forecasting}. \ans{The probabilistic intervals are added to the forecasts using 97.5-th percentile of the Student's t-distribution.}

\ans{The quality measures for medium-term (100 days) in-sample forecasts} are presented in Tab.~\ref{tab_prod_forecasting_metrics}. The root mean squared error is used to compare the forecasted oil production with the historical observations. Dynamic time warping distance (DTW) \cite{muller2007dynamic} is used to compare the forecast and observations ignoring the minor misalignment in the peaks.

\begin{table}[h!]
\resizebox{\textwidth}{!}{%
\centering
\begin{tabular}{|c|c|c|c|c|c|c|c|c|c|c|}
\hline
\makecell{Production \\ wells} & \multicolumn{10}{c|}{Production wells} \\\hline
\multirow{2}{*}{Model} & \multicolumn{2}{c|}{5351} & \multicolumn{2}{c|}{5599} & \multicolumn{2}{c|}{7078} & \multicolumn{2}{c|}{7289} & \multicolumn{2}{c|}{7405f} \\ \cline{2-11} 
 & RMSE & DTW & RMSE & DTW & RMSE & DTW & RMSE & DTW & RMSE & DTW \\ \hline
CRM & 331 & 258 & 338 & 207 & 592 & \textbf{496} & 276 & 145 & 588 & 291 \\\hline
ML & \textbf{96} & \textbf{209} & 269 & \textbf{166} & \textbf{491} & 603 & \textbf{78} & 107 & 529 & 197 \\ \hline
\makecell{Auto \\ ML+CRM}  & 108 & \textbf{210} & \textbf{184} & 187 & 493 & \textbf{593} & \textbf{79} & \textbf{105} & \textbf{408} & \textbf{179} \\
\hline
\end{tabular}}
\caption{\ans{The quality metrics for the in-sample oil production forecast obtained with different approaches: CRM, machine learning model \ansnew{based on} random forest regression with lagged features, and evolutionary optimised hybrid model (AutoML+CRM). DTW (divided by $10^{3}$) and RMSE (in $m^{3}$) measures are used (the lower values are better). The forecast length is 100 days, number of iterations is 4.}}
\label{tab_prod_forecasting_metrics}
\end{table}

It can be seen that CRM reproduces sharp production changes, but many parts of the forecast are very inaccurate. The specifics of the Volve field can explain the insufficient quality of the balance model results - the available dataset represents the late stage of its development. The ML-based forecast provides better representation, especially with the involvement of the CRM in a frame of a hybrid model. 

\ans{It can be seen that the evolutionary automation of ML pipeline design provides a better quality of short-term forecasting in many cases. However, several cases show the lower quality of auto-generated models. It can be explained by the high uncertainty caused by the small size of the dataset and the high variability of the oil production. We can conclude that additional improvement of the AutoML techniques in controllability and robustness is required.}

\subsection{Automated seismic analysis}
\label{subsec_exp_seismic}

\begin{figure}[ht!]
\centerline{\includegraphics[width=0.7\textwidth]{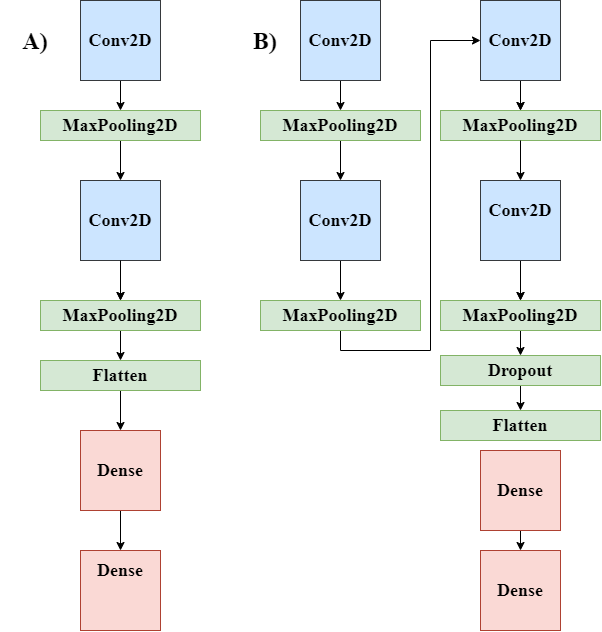}}
\caption{\ans{The architectures of the deep models used in the experimental studies: a) baseline CNN model identified by expert; b) CNN model obtained by automated NAS-based approach.}}
\label{fig_NAS_model}
\end{figure}

\ans{As mentioned in Section~\ref{sec_problem}, the presence of anomalies in the amplitudes of seismic traces of reflected waves (called "bright spots" or "dark spots") can be a sign of the presence of hydrocarbon deposits. However, it should be noted that this is only one of the many ways that geologists identify hydrocarbon deposits. In many cases, this approach can lead to false-positive prediction, as an example in cases when hydrocarbons are not detected in reservoirs in the rock. The solve this problem, we use the automatic machine learning (AutoML) approach. First, we can customize the loss function to minimize false-positive errors. Second, detecting a reservoir is usually associated with an imbalance of classes, and we can obtain a more acceptable model than in traditional machine learning approaches. Third, since the threshold value for the probability of assigning a seismic time slice to the reservoir class is a parameter, we can vary the size of the proposed reservoir \ansnew{in an automated manner} and use this information as an addition to the complex analysis of the oil field.}
As a baseline for the experiments with seismic data analysis, we used a convolution neural network implemented by the expert. The neural network architecture is shown in Fig.~\ref{fig_NAS_model}(a). 

For the seismic data analysis experiment, the search for optimal architecture is carried out over 30 epochs. The sizes of the training and validation data sets are increased to 1700 and 300 samples, respectively. The accuracy metric is selected as the loss function. The best-obtained accuracy for the training set is 97\% and 95\% for the validation set.

\begin{figure}[ht!]
\centerline{\includegraphics[width=1.0\textwidth]{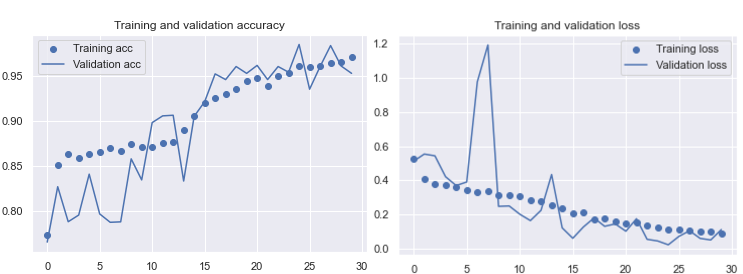}}
\caption{Dynamics of quality metrics for CNN model during training. Plot on the left shows change in value of accuracy metric for training and validation data sets. Plot on the right shows changes in value of the loss function for the same data sets. Dots and line indicate values for the training and validation data set respectively.}
\label{fig_preig_metrics_for_NAS}
\end{figure}

 Fig.~\ref{fig_NAS_model}(b) demonstrates the architecture of final NAS-model. The quality measures for the seismic classification models are presented in Tab.~\ref{tab_seismic_metrics}.  Fig.~\ref{fig_preig_metrics_for_NAS} shows graphs of changes in quality metric and loss function on training and validation data sets.
 
 As we can see in Fig.~\ref{fig_NAS_model}, the model obtained using NAS architecture has two additional convolution layers and two pooling layers as well as a dropout layer before dense layers (in comparison with the baseline model). Adding a dropout layer is a regularization of the model and helps solve the overfitting problem during the training process.

\begin{table}[]
\resizebox{\textwidth}{!}{%
\begin{tabular}{|c|c|c|c|c|c|c|c|}
\hline
\multirow{2}{*}{Model} & \multicolumn{2}{c}{Train}       & \multicolumn{5}{|c|}{Validation}    \\\cline{2-8}
                       & Accuracy       & \makecell{ROC\\AUC}        & Accuracy       & \makecell{ROC\\AUC}        & Precision      & Recall         & F1             \\
\hline

\makecell{CNN \\ baseline}           & 0.935          & 0.803          & 0.858          & 0.876          & 0.850          & \textbf{0.930} & 0.888          \\
\hline
\makecell{CNN \\ NAS}                & \textbf{0.972} & \textbf{0.907} & \textbf{0.954} & \textbf{0.953} & \textbf{0.907} & 0.907          & \textbf{0.907} \\ 
\hline
\end{tabular}}
\caption{The quality metrics for the oil reservoir detection obtained with different approaches: CNN baseline model, CNN model obtained from NAS. The accuracy, ROC AUC, precision, recall and F1 measures are used (the highest values are better).}
\label{tab_seismic_metrics}
\end{table}

These neural network architecture changes allowed us to significantly increase quality metrics on the validation dataset, while values of quality metrics on the training set will not differ so significantly. \ans{It shows that the neural model designed in an automated way} has a better generalizing ability and can produce better results on new data than the baseline model.

It is seen from Fig.~\ref{fig_preig_metrics_for_NAS} that the minimum number of epochs for training should be taken at least 20 because otherwise, the variance in values of validation accuracy and validation loss metrics are too high, which is usually a sign of an unfinished model training procedure.
The obtained model reasonably identified single reservoirs of a simple shape in the image. However, based on the predictions analysis, it may be concluded that the model does not recognize the reservoirs of more complex configurations (elongated, inhomogeneous).


\subsection{Semantic segmentation of seismic time slices}
\label{subsec_exp_seismic_segmenation}

\begin{figure}[ht!]
\centerline{\includegraphics[width=1.0\textwidth]{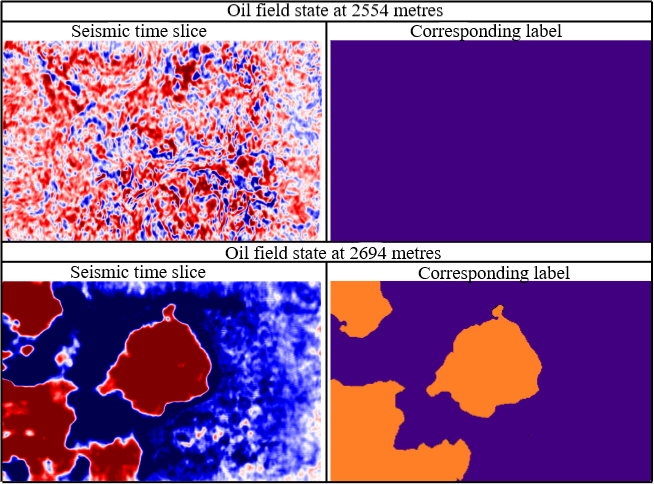}}
\caption{Samples of seismic time slices and corresponding labels. As can be seen on the seismic time slice made at a depth of 2554 meters, an expert could not identify the contact boundary between geological elements, while on the seismic time slice located at a depth of 2694 meters, the expert identified several separate geological elements at once.}
\label{fig_unet_slices}
\end{figure}

For the semantic segmentation of seismic time slices, we used 450 images manually labeled by the expert to highlight the reservoir's boundaries. The examples of the results are presented in Fig.~\ref{fig_unet_slices}. The images on the left are raw seismic time slices. The images on the right are masks representing ground truth for the segmentation. This is the result that the model should predict for a given seismic time slice. The violet area indicates the absence of an oil reservoir, and the orange area indicates the presence of an oil reservoir.

\begin{figure}[ht!]
\centerline{\includegraphics[width=1.0\textwidth]{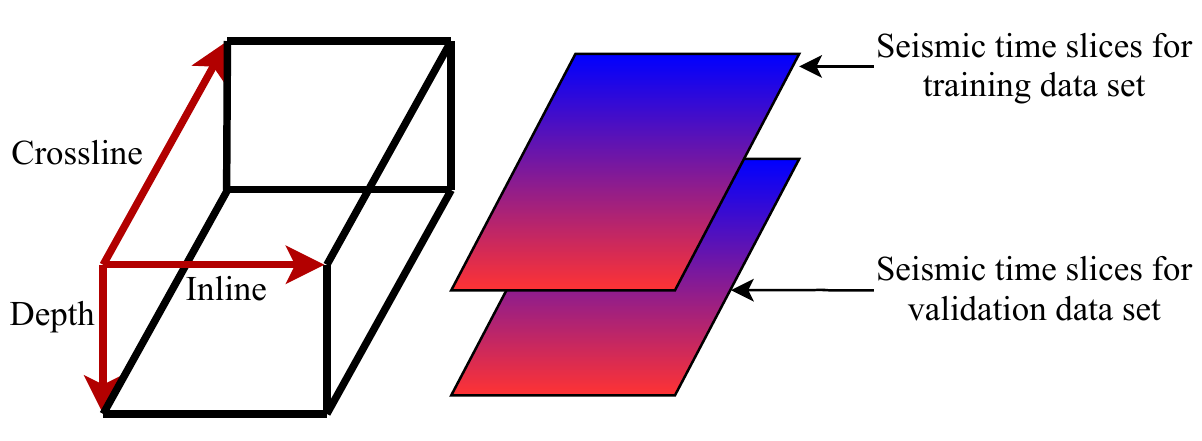}}
\caption{Scheme for the generation of training and validation samples from images of seismic sections. The main idea is that the training data set will contain seismic time slices of the chalk formation (located closer to the surface of the Volve field), and the validation data set will contain slices of the Hugin sandstone reservoir (located deeper from the surface of Volve field).}
\label{fig_unet_gt}
\end{figure}

\begin{figure}[ht!]
\centerline{\includegraphics[width=1.0\textwidth]{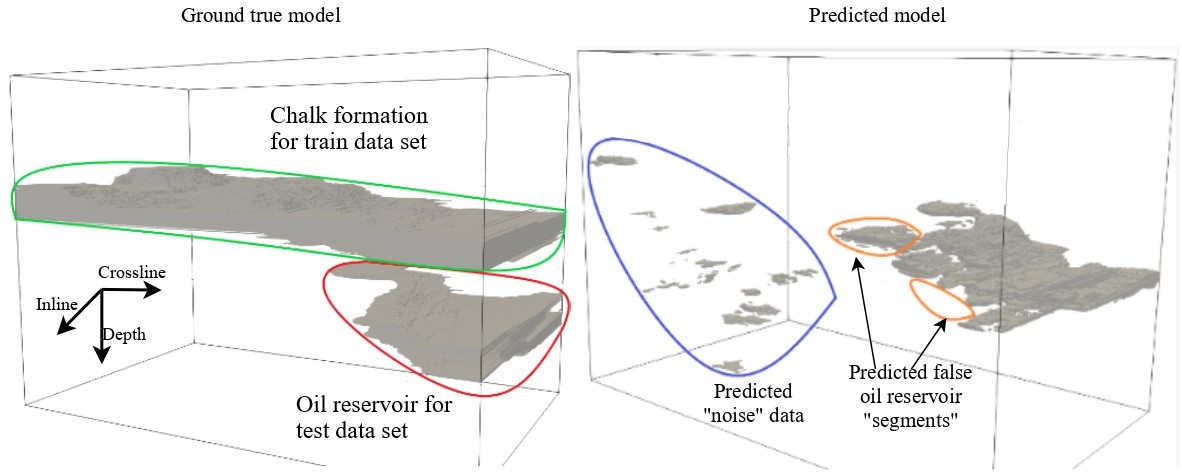}}
\caption{Three-dimensional image of oil reservoirs obtained using U-Net model. \ans{Area inside green curve is chalk formation for training data set, area inside red curve is oil reservoir for test data set.}}
\label{fig_3d_compare}
\end{figure}

\ans{For the seismic segmentation experiment, we divide the data set into spatially separated training and validation data sets: the training data set contains images of a chalk formation highlighted by the green curve on Fig.~\ref{fig_3d_compare} and the validation data set contains images of Hugin Sandstone reservoir \citep{mancinelli2020four} highlighted by the red curve.} Using the approach shown in Fig.~\ref{fig_unet_gt}, we can prevent data leaks during the training process, and we can correctly analyze the predicted oil reservoir. For cases with more diverse data sets that contain information about several reservoirs, the analogs matching procedure \cite{martin2014new} can be applied to use different reservoirs for training and validation of the model.

As a model for the experiments with semantic segmentation of seismic time slices, we used a convolutional U-Net architecture \cite{unet2015}. The model is compiled with Adam optimizer, and the batch size is set to 32. \ans{The binary cross-entropy loss function is used since there are only two states ("Potential oil reservoir" or "No oil reservoir").} 

\begin{figure}[ht!]
\centerline{\includegraphics[width=1.0\textwidth]{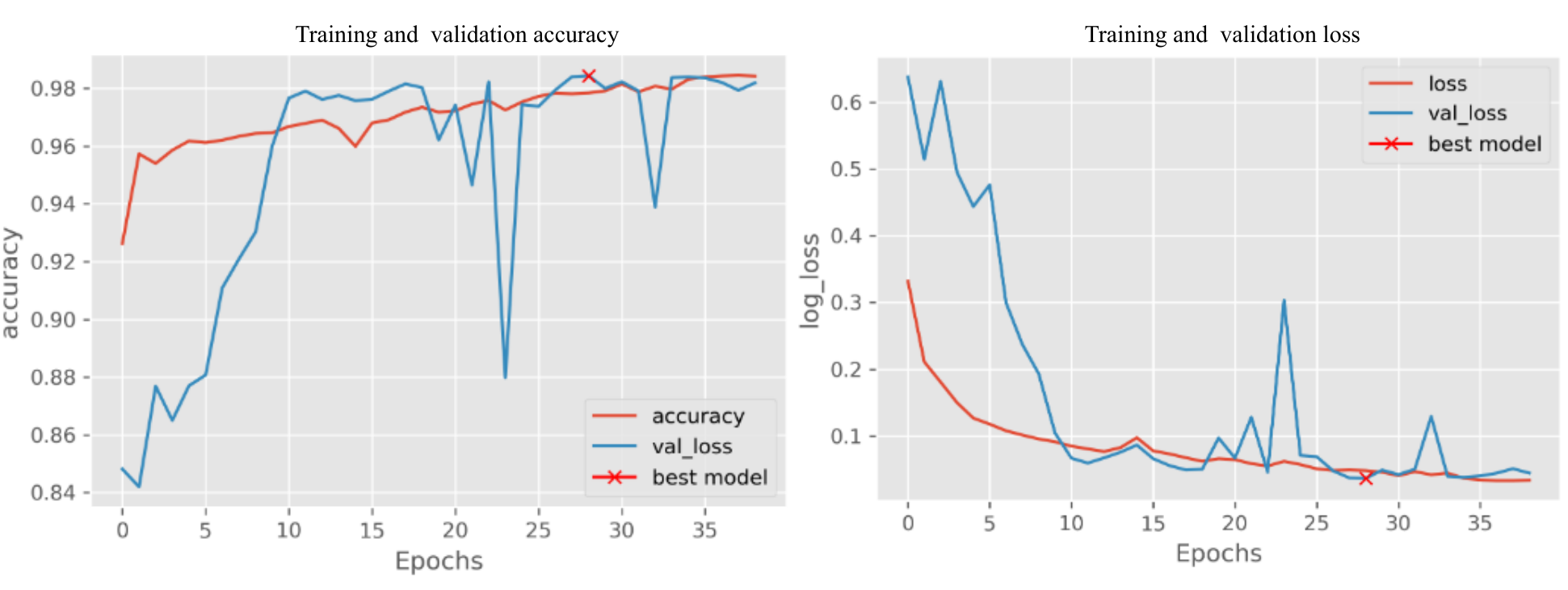}}
\caption{The changes in quality metrics for the U-Net model over training time. The red line on the graph shows the change in the value of the loss metric for the training data set. The blue line on the graph shows changes in the value of the loss metric for the validation data set. The red cross shows the epoch at which the minimum value of the loss function was reached on training and validation data sets.}
\label{fig_Unet_curve}
\end{figure}

The sizes of the training and validation data sets are increased to 1436 and 360 samples, respectively, by non-physical data augmentation. The best-obtained pixel accuracy for the training set is 98,43\% and 98,16\% for the validation set. The quality measures for the semantic segmentation of seismic time slices are presented in Fig.~\ref{fig_Unet_curve} and Tab.~\ref{tab_segmentation_metrics}. 

\begin{table}[ht!]
\centering
\begin{tabular}{|c|c|c|c|}
\hline
\ Data set & Number of samples & Accuracy & meanIoU \\
\hline
\ Train& 1436& 0.9843 & 0.8409\\
\hline
\ Validation & 360& 0.9816& 0.7931 \\ 
\hline
\end{tabular}
\caption{The quality metrics of U-Net segmentation model that obtained for training and validation data sets. The accuracy and \ansnew{meanIoU (Intersection over Union - estimation of the overlap between two areas)} measures are used.}
\label{tab_segmentation_metrics}
\end{table}

\begin{figure}[ht!]
\centerline{\includegraphics[width=1.0\textwidth]{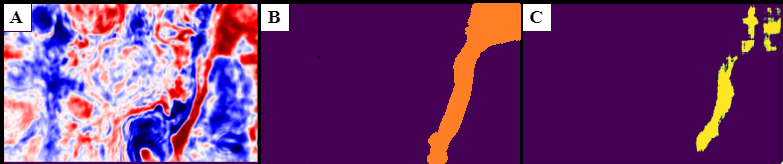}}
\caption{The examples of original slices associated with U-Net predictions: a) Seismic time slice; b) Ground true label; c) Predicted label.}
\label{fig_predictions_compare2}
\end{figure}

We get a value between 0 to 1 for every pixel. This value represents the probability of oil reservoir presence. The closer the value is to 1, the higher the probability of reservoir presence. In our case, we take 0.8 as the threshold for classifying a pixel as 0 or 1 and thus get a binary mask, represented in Fig.~\ref{fig_predictions_compare2} where orange color indicates the presence of an oil reservoir, and violet indicates the absence of it. The threshold can be considered another hyperparameter that can be optimized during the model construction process.

As it can be seen from Fig.~\ref{fig_3d_compare}, our model was able to quite reliably \ansnew{restore} the shape of the oil reservoir that was included in the test set. It should be noted that the errors in the predicted results (the area inside the blue curve is noise in the data, the area inside the orange curves is incorrect predicted segments of the oil reservoir in the test sample) are not critical for the correct interpretation. The predicted oil reservoir's parameters (shape, size, location) accurately match the "ground truth" parameters.

The results of the application of the method for interpreting seismic time slices using the U-Net model demonstrate that the proposed approach can perform segmentation that is close to the human expert interpretation results. This method allows to speed up the processing and analysis of seismic survey data and simplify the interpretation process for the domain experts.

\section{Conclusion}
\label{sec_concl}

\ans{In the paper, we describe the usage of automated data-driven approaches for tasks of oil field development analysis}. Two sub-tasks were formulated and solved: oil production prediction (time series forecasting problem) and seismic-based reservoir detection (pattern recognition problem). The Volve field open dataset was used to conduct the experimental studies. 

The obtained results confirm that the automated hybrid approaches that involve the machine learning models into the composite pipelines allow increasing the quality of the modeling and analysis compared with the physics-related model. Also, the application of automated modeling tools makes it possible to decrease the expert's involvement in the model structure development and increase the quality of the obtained results. 

During the future development of this study, more complex neural models can be used for production forecasting, and the seismic analysis problem can be formulated more comprehensively (e.g., the prediction of specific reservoir characteristics).

\section*{Authorship contribution statement}

Nikolay Nikitin: Methodology, Software, Visualization, Writing – original draft. Ilia Revin: Software, Visualization, Writing – original draft. Alexander Hvatov: Methodology, Writing – original draft. Pavel Vychuzhanin: Software, Validation. Anna Kalyuzhnaya: Conceptualization, Project administration

\section*{Acknowledgements}

This research is financially supported by the Ministry of Science and Higher Education, Agreement FSER-2021-0012.

\section*{Code and Data Availability}

The software that implements the described approaches is named ML-oil-field-modelling-tool and available in the GitHub repository
(\url{https://github.com/ITMO-NSS-team/oil-field-modelling}) under the open-source license BSD-3. The developers are Nikolay Nikitin, Ilia Revin, and Pavel Vychuzhanin.

The publication year is 2020. The required additional software is Python 3.8 interpreter and the libraries described in requirements.txt. Also, the Jupyter Server should be installed to run the examples implemented as Jupyter Notebooks.


The self-developed FEDOT framework that was used in the experiments is also available in the GitHub repository
(\url{https://github.com/nccr-itmo/FEDOT}) under the open-source license BSD-3.

The raw Volve field data are available at \url{https://www.equinor.com/en/how-and-why/digitalisation-in-our-dna/volve-field-data-village-download.html}. The processed data are located in \url{https://data.mendeley.com/datasets/g2vxy237h5/1}.






\bibliographystyle{elsarticle-num-names}
\bibliography{ref.bib}







\end{document}